# Towards Unsupervised Learning of Temporal Relations between Events

**Seyed Abolghasem Mirroshandel**    MIRROSHANDEL@CE.SHARIF.EDU
**Gholamreza Ghassem-Sani**    SANI@SHARIF.EDU
*Computer Engineering Department*
*Sharif University of Technology*
*Azadi Avenue, Tehran 11155-9517, Iran*

## Abstract

Automatic extraction of temporal relations between event pairs is an important task for several natural language processing applications such as Question Answering, Information Extraction, and Summarization. Since most existing methods are supervised and require large corpora, which for many languages do not exist, we have concentrated our efforts to reduce the need for annotated data as much as possible. This paper presents two different algorithms towards this goal. The first algorithm is a weakly supervised machine learning approach for classification of temporal relations between events. In the first stage, the algorithm learns a general classifier from an annotated corpus. Then, inspired by the hypothesis of "one type of temporal relation per discourse", it extracts useful information from a cluster of topically related documents. We show that by combining the global information of such a cluster with local decisions of a general classifier, a bootstrapping cross-document classifier can be built to extract temporal relations between events. Our experiments show that without any additional annotated data, the accuracy of the proposed algorithm is higher than that of several previous successful systems. The second proposed method for temporal relation extraction is based on the expectation maximization (EM) algorithm. Within EM, we used different techniques such as a greedy best-first search and integer linear programming for temporal inconsistency removal. We think that the experimental results of our EM based algorithm, as a first step toward a fully unsupervised temporal relation extraction method, is encouraging.

## 1. Introduction

Much progress has been made in natural language processing (NLP) in recent years. Combining statistical and symbolic methods has played a significant role in these advances. As a result, tasks such as part-of-speech tagging (Søgaard, 2011), parsing (Petrov & Klein, 2007), and named entity recognition (Mikheev, Grover, & Moens, 1998) have been addressed with satisfactory results. However, in some other tasks such as temporal information processing, which need a deeper analysis of meaning, the achieved results have not yet been as satisfactory.

Temporal information is encoded in the textual description of events. Lately, the increasing attention to practical NLP applications such as question answering, summarization, and information extraction have resulted in a growing demand of temporal information processing (Tatu & Srikanth, 2008). In question answering, one may expect the system to answer questions such as *"when an event occurred"*, or *"what is the chronological order be-* *tween some desired events"*. In text summarization, especially in the multi-document type, knowing the order of events is a useful source for correctly merging related information.

Construction of the TimeBank corpus in 2003 (Pustejovsky et al., 2003), provided the opportunity of applying different machine learning methods to the task of temporal relation extraction. However, it has been realized that even a six-class classification of temporal relations is a very complicated task, even for human annotators (Mani, Verhagen, Wellner, Lee, & Pustejovsky, 2006).

This paper presents two different approaches in which the need for annotated data in temporal relation learning is reduced. The first approach is a weakly supervised machine learning algorithm for classification of temporal relations between events. In the first stage, the algorithm learns a general classifier from an annotated corpus. Then, inspired by the hypothesis of "one type of temporal relation per discourse", it extracts useful information from a cluster of topically related documents for retraining of the model. By combining the global information of such a cluster with local decisions of a general classifier, we propose a novel bootstrapping cross-document classifier to extract temporal relations between events. Our experiments show that without any additional annotated data, the accuracy of the proposed algorithm is at least 7% higher than that of the state-of-the-art of statistical methods (Chambers, Wang, & Jurafsky, 2007).

The second introduced approach is a novel usage of expectation maximization (EM) algorithm for temporal relation learning. This algorithm also employs Allen's interval algebra (Allen, 1984) for correction of predicted relations. For applying interval algebra, we utilize two different approaches: 1) a heuristic search method and 2) integer linear programming (ILP). We think that the experimental results of this EM based algorithm, as a first step toward a fully unsupervised temporal relation extraction method, is encouraging.

The remainder of this paper is organized as follows: section 2 is about previous approaches to the temporal relation learning. Section 3 explains our first proposed method, which is evaluated in section 4. The second algorithm is explained in section 5, and evaluated in section 6. Finally, section 7 includes our conclusions and some possible future work.

## 2. Temporal Relation Learning

Assuming that we have access to the texts in which events and time expressions have been appropriately tagged, two different tasks pertaining to temporal relation learning can be distinguished: 1) detecting whether there exist any relation between a given pair of events/time expressions; 2) identifying the relation type for positive cases of the first task. The first task is very hard to evaluate, because the annotators may ignore many plausible existing relations while tagging the corpora (Mani et al., 2006). Accordingly, in this paper like other existing research, we have only addressed the second task, which can be more specifically defined as follows: *"For a given ordered pair of components $(x_1, x_2)$, where $x_1$ and $x_2$ are annotated events and/or time expressions, a temporal relation classifier identifies the type of relation $r_i$ that temporally links $x_1$ to $x_2$"*. As it is shown in Figure 1, each temporal relation can be one of the fourteen types proposed in TimeML (Pustejovsky et al., 2003). For example, in *"Powerful political <u>pressures</u> (event$_1$) may <u>convince</u> (event$_2$) the Conservative government to <u>keep</u> (event$_3$) its so-called golden share, which limits any individual holding to* 15%, until the <u>restriction</u> (event4) <u>expires</u> (event5) on <u>Dec. 31, 1990</u> (time1)". (taken from document *wsj_0745* of TimeBank, see Pustejovsky et al., 2003). The task is to automatically tag the relations between pairs $(event_1, event_2)$, $(event_3, event_5)$, $(event_5, event_4)$, and $(event_5, time_1)$ with *BEFORE, ENDED_BY, ENDS*, and *IS_INCLUDED*, respectively (see Figure 2). Since automatic extraction of just *"event-event"* relations is itself a difficult task, in this paper, we have focused on this particular task, and left the detection of other type of temporal relations such as *"event-time"* or *"time-time"* to future work.

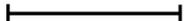

Figure 1: Different temporal relations in TimeML.

There are many ongoing research focusing on temporal relation learning. Additionally, there have been two important shared tasks on temporal information extraction: TempEval 2007 (Verhagen et al., 2007) and TempEval 2010 (Verhagen, Sauri, Caselli, & Pustejovsky, 2010). In TempEval 2007, there were three different tasks regarding temporal relations classification between A) events and times within the same sentence; B) creation time of a document and its events; and C) main (verb) events in adjacent sentences.

In TempEval 2010, there were six different tasks including A, B) Determining the time expressions and events of input texts and specified features; and temporal relation classification between C) events and times within the same sentence; D) creation time of a document and its events; E) main events in consecutive sentences; and F) two events where one event syntactically dominates the other event.

Due to focusing on temporal relations between event pairs, task C of TempEval 2007 plus tasks E and F of TempEval 2010 are similar to the task that we tackle in this paper; however, these tasks can be considered as special cases of ours. For instance, in task E of TempEval 2010, only the event pairs from consecutive sentences are considered; whereas, in our task the event pairs can be either from the same sentence or from any other two sentences of the input text.

The research on temporal relation learning can be divided into different categories. In this paper, we divide these efforts into three groups: 1) Statistical; 2) Rule-based, and 3)

Hybrid, which are explained in the following sections.

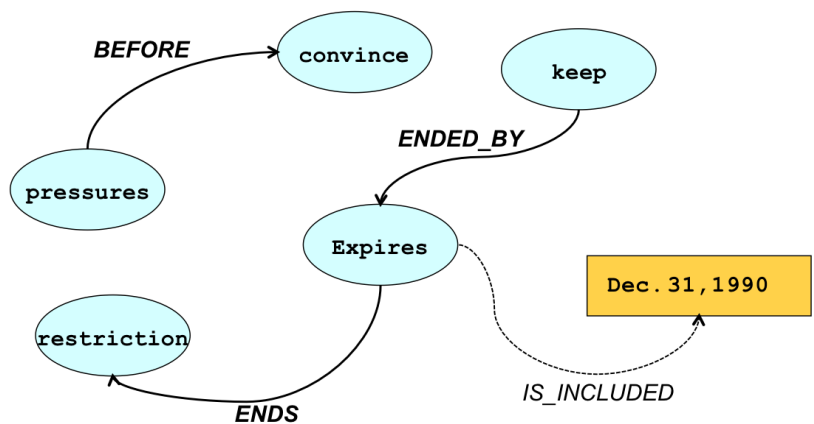

Figure 2: Temporal relations of the sentence *"Powerful political pressures (event₁) may convince (event₂) the Conservative government to keep (event₃) its so-called golden share, which limits any individual holding to 15%, until the restriction (event₄) expires (event₅) on Dec. 31, 1990 (time₁)"*. Bold arrows show relations between event pairs.

## 2.1 Statistical Methods

In all statistical methods, a classification (or clustering) algorithm is employed over a number of tagged and/or extracted features of an input corpus. Maximum Entropy (Mani, Wellner, Verhagen, & Pustejovsky, 2007; Derczynski & Gaizauskas, 2010), Support Vector Machines (Chambers et al., 2007; Bethard & Martin, 2007; Hepple, Setzer, & Gaizauskas, 2007; Cheng, Asahara, & Matsumoto, 2007; Mirroshandel, Ghassem-Sani, & Khayyamian, 2009a, 2009b, 2011), Conditional Random Fields (Llorens, Saquete, & Navarro, 2010; Kolya, Ekbal, & Bandyopadhyay, 2010), and Markov Logic Networks (UzZaman & Allen, 2010; Ha, Baikadi, Licata, & Lester, 2010) are some of the statistical techniques that have been applied to this problem.

MaxEnt is one of the first approaches to the temporal relation learning, which uses maximum entropy classification algorithm (Mani et al., 2007). In this method, the classifier assigns one of six different temporal relation types to each event-event or event-time pair. The classifier relies on a number of features including modality, polarity, tense, aspect, and the event class, which have been hand-tagged in the corpus. In addition to these features, it also relies on pairwise agreement of two additional features: tense and aspect. We later propose a new technique to improve MaxEnt. The results of comparing the proposed method and MaxEnt are given in section 4.

USFD2 (Derczynski & Gaizauskas, 2010) is another method that employs maximum entropy in solving tasks C and F of TempEval 2010. This method uses the same features as MaxEnt plus some features related to the so-called signals of the text. USFD2 achieved the second highest score in task C of TempEval 2010. However, its results on task F were not satisfactory enough.

The state-of-the-art of the statistical methods is analogous to MaxEnt (Chambers et al., 2007). It works in two consecutive stages and employs some event-event features in addition to those used by MaxEnt. In this work, Support Vector Machines (SVM) are used for classification. Similar results were reported for using a Naive Bayes classier instead of SVM. Section 4 also includes the results of comparing this work with our proposed algorithm.

SVMs have been also used as a classification algorithm in several other research. CU-TMP (Bethard & Martin, 2007) applied SVM for solving all three tasks of TempEval 2007. It also used gold-standard TimeBank features for event and time expressions plus parts of derived parse trees from the input text. CU-TMP first solves task B and then uses its results to tackle tasks A and C.

USFD (Hepple et al., 2007) and NAIST-Japan (Cheng et al., 2007) were two other participants of TempEval 2007 that used SVM for classification. In NAIST-Japan, the task was defined as a sequence labeling model. The task is approached by using HMM-SVM, relying on features from dependency-parsed trees and standard attributes of target events/time expressions. The result of this system was slightly more than average in tasks A and B, but less than average in task C. It was shown that extracted features from dependency parsed trees were not so effective for task C, in contrast with tasks A and B. In USFD, temporal relation learning is treated as a simple classification task (Hepple et al., 2007). They used different classification algorithm from WEKA machine learning workbench (Hall et al., 2009). In task C, SVMs have gained the best result among the participants of the shared task.

In another work, a corpus of parallel temporal and causal relations was employed, and SVMs were used to extract both types of relations (Bethard & Martin, 2008). Since existing corpora provide no parallel temporal or causal annotations, 1000 conjoined event pairs were annotated (Bethard & Martin, 2008; Bethard, 2007). It was shown that causal relation information could be helpful in temporal relation extraction, too. It was also shown that temporal relation information has mutually positive effects in causal relation extraction (Bethard & Martin, 2008).

Bethard and his colleagues (2007a, 2007b) have applied SVM to classify event pairs in which the first event is a verb and the second one is the head of a clausal argument of that verb. They have used a combination of a number of event based features (e.g., tense and aspect) and some syntactic features (e.g., a specific path through the parse tree). Their reported results have shown a high accuracy for these specific event pairs.

There are also other algorithms which utilize grammatical information in SVM using convolution tree kernels (Mirroshandel et al., 2009a, 2009b, 2011). It was shown that grammatical aspects of the input text are rich sources of information for temporal relation classification. Argument Ancestor Path Distance (AAPD) convolution tree kernel is the most successful tree kernel that has been used in SVM classification. This kernel is similar to the CollinsDuffy tree kernel (Collins & Duffy, 2001). The CollinsDuffy kernel effectively counts the number of common subtrees between two comparing parse trees. In this kernel, all subtrees have the same importance, whereas in AAPD, different weighting functions are used to compute the kernel value. Furthermore, in AAPD, the significance of subtrees is measured using the distance from a so-called argument ancestor path (AAP). An AAP is the ancestor nodes of an argument (event). An example of a node (NN) and the distance between the node and an AAP is shown in Figure 3. In AAPD, the closer a node is to the path, the less it is decayed by the weighting function. In other words, the nodes which are located nearer to the path are more important than those farther away.

To improve the accuracy of AAPD, it was combined with some other kernels, which were either linear or polynomial (Zhang, Zhang, Su, & Zhou, 2006). However, polynomial composite kernels have shown superior results (Mirroshandel et al., 2009b). Section 4 includes the results of comparing AAPD and AAPD Polynomial kernels.

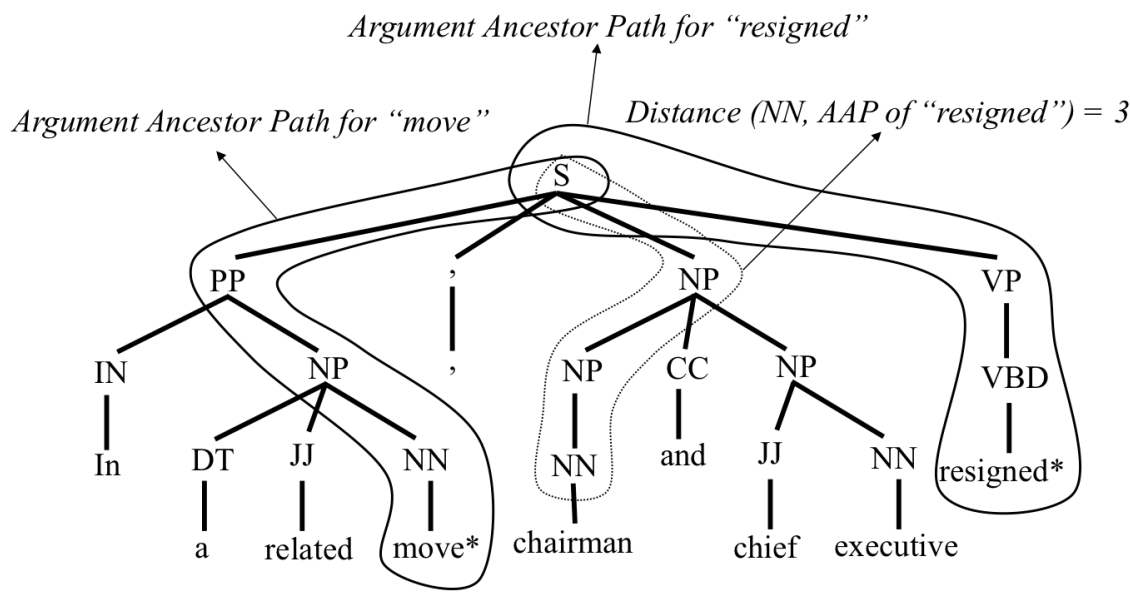

Figure 3: A syntactic parse tree with two argument ancestor paths for events *"move"* and *"resigned"* plus the distance of node *NN* from AAP of *"resigned"*.

Markov Logic Networks (MLN) is another classification algorithm, which have been used by two participants of TempEval 2010: TRIPS & TRIOS (UzZaman & Allen, 2010) and NCSU (Ha et al., 2010). TRIPS and TRIOS use a number of features produced by a deep semantic parser, plus a few features extracted from target pairs (i.e., event/time expression). In contrast with other participants, TRIPS and TRIOS operate on raw texts. In other words, these systems do not use any tagged events/time expressions. They both outperformed all other teams on two tasks (C and E). TRIOS also gained the second best results on the four remaining tasks. NCSU is another participant of TempEval 2010 that uses MLN for classification. It relies on basic annotated features, syntactic features extracted from generated parse trees, and lexical semantic features from two external resources (Ver-

bOcean and WordNet) (Ha et al., 2010). NCSU was applied to tasks C, D, E, and F in two different settings: NCSU-indi and NCSU-joint. In NCSU-indi, an independent MLN was trained for each task. On the other hand, a set of global formulae was also added to NCSU-joint to ensure the consistency among classification decisions from four local MLNs (one for each task). NCSU-indi achieved the best result in task F and the second best result on task C.

One of the most successful participants of TempEval 2010 was TIPSem that is based on Conditional Random Field (CRF) models for classification purpose (Llorens et al., 2010). TIPSem employs different morphological, syntactic, and semantic features for building CRF models. In Spanish, it achieved the best results in all tasks. In English, TIPSem achieved the best results in Tasks B and D; and was one of the best systems in all other tasks. JU_CSE_TEMP was another participant of TempEval 2010 that utilized CRF models for temporal relation learning tasks (Kolya et al., 2010). The system needs only the gold-standard features of TimeBank for time expressions and/or events. In comparison with TIPSem, JU_CSE_TEMP achieved weaker results, which shows the importance of feature engineering in temporal relation learning.

There is another approach that applies different machine learning techniques to detect intra-sentential events, and builds a corpus of sentences with two or more events in which at least one event is triggered by a key time word (e.g., after, before, etc.). The classifier is based on a number of syntactic and clausal ordering features (Lapata & Lascarides, 2006; Bramsen, Deshpande, Lee, & Barzilay, 2006).

There exist a comprehensive study about statistical methods, which compares three different interval based algebras in terms of classification accuracy, performance, and expressiveness power (Denis & Muller, 2010). There are also a few algorithms that exclusively work on temporal relation classification between events and time expressions. One of such algorithms employs cascaded finite-state grammars (for temporal expression analysis, shallow syntactic parsing, and feature generation) together with a machine learning component capable of effectively using large amounts of unannotated data (Boguraev & Ando, 2005).

There is a group of statistical methods that rely on information of argument fillers (called anchors) of every event expression as a valuable clue for recognizing temporal relations. In these methods, by looking at a set of event expressions whose argument fillers have a similar distribution, analogous event expressions are recognized. Algorithms such as DIRT (Lin & Pantel, 2001), TE/ASE (Szpektor, Tanev, Dagan, & Coppola, 2004), and that of the Pekar's system (2006) are examples of this type of statistical method.

DIRT is an unsupervised method based on an extended version of the so-called distributional hypothesis (Lin & Pantel, 2001). According to this hypothesis, words that occur in the same contexts are usually similar. Here, instead of words, the algorithm applies the distributional hypothesis to certain paths of the dependency trees of a parsed corpus.

TE/ASE, too, is an unsupervised algorithm, which has two major phases. In the first phase (called Anchor Set Extraction), the algorithm extracts similar anchors. Then, in the second phase (called Template Extraction), the system extracts templates from the resulting anchor sets. In the final part of the algorithm, some post-processing transformations are applied to the extracted templates to remove inappropriate templates (Szpektor et al., 2004).

In Pekar's approach (2006), co-occurrence of two verbs inside a locally coherent text is used to extract some useful information. This method has three major steps. First, based on the local discourse, it identifies several pairs of clauses as being related. Next, based on those related clauses, it tries to create a number of templates of verb pairs by using information such as syntactic behavior. In the last step, the algorithm scores and employs those templates for relation extraction.

## 2.2 Rule-Based Methods

The common idea behind rule-based methods is to find some general patterns for classifying temporal relations. In most of these works, rules (patterns) are manually defined.

Perhaps the simplest rule-based method is the one that was developed using a knowledge resource called VerbOcean (Chklovski & Pantel, 2005). VerbOcean has a small number of manually designed generic rules. The style of rules is in the form of "*<Verb-X> *<Verb-Y> *". For example, there are rules such as *"to Verb-X and then Verb-Y"*, *"to Verb-X and eventually Verb-Y"*, or *"to Verb-X and later Verb-Y"* for the *"happens-before"* relation type; and also there are rules such as *"Verb-X even Verb-Y"* or *"Verb-Y or at least Verb-X"* for the so-called *"strength"* relation type. After manually creating these rules, a number of semantic relations (e.g., *strength, antonymy, happens-before,* etc.) between events can be detected. Several heuristics were also employed to filter inappropriate relations (Chklovski & Pantel, 2005).

There is another rule-based method for temporal relation learning focused on biomedical texts (Mulkar-Mehta, Hobbs, Liu, & Zhou, 2009). It was shown that existing methods for temporal relation learning were not effective for such texts. In this work, some specific axioms (rules) were used to predict the temporal and causal relations. A pattern extraction algorithm was employed to create the system rules in a semi-automatic manner.

XRCE-T, a participant of TempEval 2007, is a rule-based system that relies on syntactic and semantic features (e.g., deep syntactic analysis and determination of thematic roles) (Hagège & Tannier, 2007). XRCE-T was in fact used as a post-processing module of a general purpose linguistic analyzer.

In another study, rules of temporal transitivity were used to increase the training set. The test accuracy on this enlarged corpus showed some improvements (Mani et al., 2007).

Reasoning with pre-determined rules is another approach to the rules' usage. In the work of Tatu and Srikanth (2008), a rich set of axioms (rules) was created and used by a first order logic based theorem prover to find a proof for each temporal relation by refutation.

A set of discourse rules was used in the algorithm of Muller and Tannier (2004) to establish the possible relations between every two consecutive events of the input text. These rules were based on tenses of the event verbs. Then a classical path-consistency algorithm (Allen, 1984) was applied to the extracted relations of the first step.

## 2.3 Hybrid Methods

It has been shown that one can increase the accuracy of temporal relation classifiers by merging some discussed methods. For example in the work of Chambers and Jurafsky (2008), local decisions generated by a statistical method were combined with two types of implicit global rule-based properties. These properties included the transitivity rule (e.g., A *before* B and B *before* C implies A *before* C), and time expressions normalization (e.g., last month is *before* yesterday). The constraints were used to create a more densely-connected network of events, and then a global state of consistency was enforced by incorporating the constraints into an integer linear programming framework (Chambers & Jurafsky, 2008).

Integer linear programming with local classifiers was shown to be appropriate only for cases in which the number of possible relations between events is restricted (Denis & Muller, 2011). It was suggested that a translation of constraints from temporal intervals to their endpoints can be used to handle a significantly smaller set of constraints. During translation, temporal relations are preserved. This method was shown to have a rather high accuracy. They also proposed a graph decomposition technique that can further improve the accuracy.

In another algorithm, which was spiritually similar to that of (Chambers & Jurafsky, 2008), instead of applying global constraints using integer linear programming, the so-called Markov logic (ML) was used (Yoshikawa, Riedel, Asahara, & Matsumoto, 2009). Global constraints can be easily captured through adding some weighted first order logic formulas. It was shown that the problem can be solved by ML more easily and accurately than by ILP.

WVALI is another hybrid system, which has an enhanced classification process by using some rules from a particular knowledge base (Puscasu, 2007). In this system, different heuristics and temporal reasoning mechanism have been combined with statistical data extracted from the training corpus. WVALI achieved the best results in all tasks of TempEval 2007.

LCC-TE was another hybrid system of TempEval 2007. It combined different machine learning models with human rules for temporal relation learning (Min, Srikanth, & Fowler, 2007). LCC-TE uses gold-standard features available in TimeBank, as well as a number of derived and extended features such as grammatical and semantic features. The evaluations on LCC-TE have shown acceptable results in all three tasks.

## 3. Bootstrapped Cross-Document Classification (BCDC)

In this section, a new method of extracting temporal relations between events is introduced. We call this method Bootstrapped Cross-Document Classification (BCDC). The results of experiments with BCDC show a significant improvement over previous work in terms of accuracy (see Tables 6 and 7). We have used SVM with three different kernels in the learning process. There are two novelties in our bootstrapping (self-training) method: 1) it is an information retrieval based approach that extracts useful information exclusively from related documents. 2) It builds a specific model for each test document. Before describing BCDC, our motivation is briefly explained in the next section.

### 3.1 Motivation

In a regular corpus with heterogeneous documents, verbs, which often act as event triggers, may have different senses in different documents. For example, event *"firing"* may have a sense of *"shooting"* a gun in a document about army, whereas it may also have a sense of *"ending"* someone's job in a different document about a company. However, for a cluster of topically-related documents, the distribution should be much less divergent. This motivated us to apply the so-called *"one sense per discourse"* hypothesis (Yarowsky, 1995) to the problem of temporal relation classification, and extend the scope of discourse from a single document to a cluster of topically related documents. Also inspired by another work that proposed assumptions of one event trigger sense and one event argument role per discourse (Ji & Grishman, 2008), we based our work on an analogous assumption, which we called *"one type of temporal relation per discourse"*. In other words, we assume that similar event pairs in different places of topically related documents are very likely to have the same temporal relations. Although, as it is later explained, we have not explicitly employed this assumption in our proposed algorithm, we have tried to verify the assumption by considering temporal relations of the Opinion corpus (Mani et al., 2006). In this corpus, documents are located in four different directories each having a specific topic. In our verification, we have considered all documents within the same directory as being related. In other words, two documents are considered as "related documents" if they are in the same directory (i.e., they have the same topic). To verify the assumption, we selected those event pairs that have appeared more than once. In Opinion corpus, there are a total number of 2666 temporally related event pairs (i.e., TLinks), out of which, only 994 pairs appeared more than once[1]. Table 1 shows the results of our verification. Supporting samples are those event pairs that have appeared in two or more related documents with exact the same temporal relation. Even if event pairs having different relations are from unrelated documents, they are also regarded as the supporting samples. On the contrary, if event pairs having different relations are from related documents, they are considered as contradictory samples. As it is shown in Table 1, more than 95% of the samples have supported our assumption (i.e., "one temporal relation per discourse").

|  | Count |
|---|---|
| Supporting Samples | 942(95%) |
| Contradictory Samples | 52(5%) |
| Total | 994 |

Table 1: The distribution for assumption of *"one type of temporal relation per discourse"* over the Opinion corpus.

As an example, in the following sentences, which have been taken from different documents of the same topic (i.e., *Kenya Tanzania Embassy bombings*), the event pair (*blast* and *kill*) has *IBEFORE* temporal relation in all sentences:

*"Reports reaching here said a massive* **blast** *damaged the U.S. embassy in Nairobi ,* **killing** *40 people while wounding at least 1,000 people."*

*"More than 100 people have been* **killed** *and more than 1,000 others wounded in the* **blasts** *next to the U.S. embassies in Kenya and Tanzania on Friday."*

*"In Dar es Salaam , she laid a wreath next to the crater left by the embassy* **blast** *that* **killed** *10 people."*

---

1. Before counting the number of event pairs, we applied a lemmatizer to event words.

## 3.2 Feature Engineering

In BCDC, two types of features are used: basic and extra event-event features. Basic features are simple features related to individual events and extra event-event features are those extracted from two related events. In the next two sections, these features are explained in more detail.

### 3.2.1 Basic Features

These are simple features extracted from events. For each event, there are five temporal attributes, which are tagged in standard corpora: 1) tense; 2) grammatical aspect; 3) modality; 4) polarity, and 5) event class. Tense and aspect define temporal location and event structure; thus, they are necessary in any method of temporal relation extraction. Modality and polarity specify non-occurring or hypothetical situations. The event class shows the type of event. The range of values for these attributes is based on the work of Pustejovsky et al. (2003), and is shown in Table 2. These attributes are either annotated in the input corpus or can be automatically extracted by existing tools.

| Attribute | Range of values |
|-----------|-----------------|
| **Tense** | none, present, past, future |
| **Aspect** | none, prog, perfect, prog_perfect |
| **Modality** | none, to, should, would, could, can, might |
| **Polarity** | positive, negative |
| **Event Class** | report, aspectual, state, I_state, I_action, perception, occurrence |

Table 2: The range of values for five temporal attributes.

In addition to the five mentioned attributes, BCDC also employs the string of words that constitute each event, their part of speech tags as well as a number of contextual features including pairwise agreement of tenses and aspects. Part of speech tags of events are again either annotated in the corpora or can be determined by existing POS taggers. For example, in sentence *"He **succeeds** James A. Taylor, who ..."*, *"succeeds"* is an event with the following features:

[tense: present], [aspect: none], [modality: none], [polarity: positive], [event class: aspectual], [word: succeeds], [pos: verb]

### 3.2.2 Extra Event-Event Features

Extra event-event features are based on two related events and are automatically extracted from the input text. In our case, there are three types of these features, defined as follows:

**Event-Event parse tree**: if both events are in the same sentence, the algorithm can use the parse tree of the sentence to learn some useful syntactic properties such as domination. In a parse tree, event A dominates event B, if A is an ancestor of B. These properties are not explicit features, but rather implicit properties which can be extracted and learned by the SVM using appropriate tree kernels such as those proposed in the work of Mirroshan-del et al. (2009b). Parse trees can be extracted by a statistical parser and there is no need for any Treebank.

**Prepositional phrase**: a preposition head is often an indicator of a temporal class. Thus, we can use a new feature that indicates if an event is a part of a prepositional phrase. This information can also be extracted from the parse trees. For example, in sentence *"I saw him before the earthquake"*, the relation between events *"saw"* and *"earthquake"* can be easily determined by the word *"before"* in the prepositional phrase *"before the earthquake"*.

**Event-Event distance**: it is based on the idea that the strength of the relationship between two events is inversely related to the textual distance of those events. It means that the relationship becomes weaker as the distance increases (and vice versa). Accordingly, intra- and inter-sentential events should be treated differently. We train two separate models: one for the intra-sentential events and one for the inter-sentential ones.

### 3.3 Proposed Algorithm

BCDC applies a novel usage of bootstrapping to the classification of temporal relations between events. It works in two main stages. In the first stage, using a standard corpus, a general model is learned. Then in stage two, the general model is retrained for each test document based on some related information. Figure 4 shows the flowchart of the proposed algorithm, which is described in more detail in the following sections.

#### 3.3.1 Stage One

In stage one, BCDC employs the discussed features extracted from a standard corpus to train a general model for classification using SVM. At the end of this stage, we will have a model for temporal relation classification. However, such models, which have also been proposed before by other researchers, all have the problem of being too general. In other words, such a model does not have any specific information about the particular domain under consideration. To better deal with this problem, BCDC has an extra bootstrapping phase of training.

#### 3.3.2 Stage Two

In stage two, we retrain the general model produced in stage one, for each test document with some related information. In order to achieve this goal, the bootstrapping phase of BCDC proceeds according to the following steps:

**Step 1**: first of all, an unprocessed test document is randomly selected.

**Step 2**: then BCDC finds the top N documents that are topically related to the selected test document from a large unannotated corpus. The choice of related documents can be made by the INDRI retrieval system (Strohman, Metzler, Turtle, & Croft, 2005). Note that the mentioned large unannotated corpus is different from the training and test corpora.

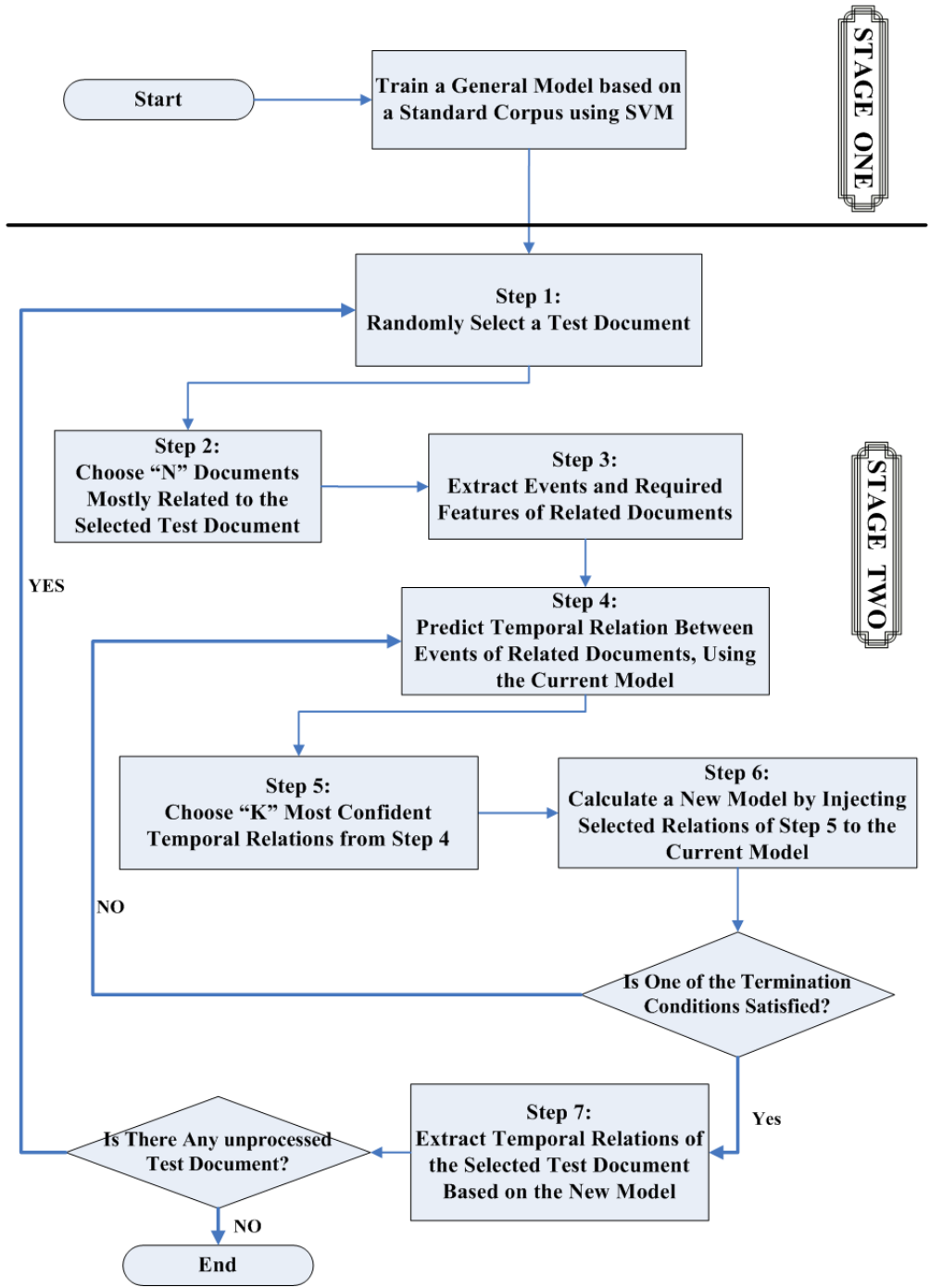

Figure 4: The flowchart of BCDC.

**Step 3**: in this step, we extract events and required features from the related documents found by INDRI. The events and specified features of section 3.2 can be automatically annotated by EVITA (Saurí, Knippen, Verhagen, & Pustejovsky, 2005). Although some of the events and/or features extracted by EVITA may be incorrect, our experimental results show that they can still be very helpful. The extra event-event features and other required features can be extracted by a POS tagger and a statistical parser.

**Step 4**: then by using the existing model, the temporal relations between only intra-sentential event pairs of the related documents are predicted. Besides, a normalized measure of confidence is computed for each relation. We have used SVM for our classification purpose. Therefore, we have designed a confidence measure using SVM, as it is explained below.

In SVM binary classification, positive and negative instances are linearly partitioned by a hyper-plane (with maximum marginal distance to instances) in the original or a higher dimensional feature space. In order to classify a new instance $X$, its distance to the hyper-plane is computed and $X$ is assigned to the class that corresponds to the sign of the computed distance. The distance between instance $X$ and hyper-plane $H$, which can be either a positive or a negative value, is supported by the support vectors $X_1 \ldots X_l$ and computed by equation 1 (Han & Kambert, 2006):

$$d(X, H) = \sum_{i=1}^{l} y_i \ \alpha_i \ X_i \ X^T \ + \ b_0 \tag{1}$$

where $y_i$ is the class label of support vector $X_k$; $\alpha_k$ and $b_0$ are numeric parameters that are automatically determined.

We have used one-versus-one case of multi-class classification with $m$ classes, in which a set of $m * (m - 1) \ / \ 2$ hyper-planes (i.e., one hyper-plane for every class pair) denoted by $H$ is defined. The hyper-plane that separates class $i$ and $j$ is referred to as $H_{i,j}$. $\mathcal{H}_i$ is used to denote a subset of $m - 1$ hyper-planes of $H$ that separates class $i$ from the others. In order to classify a new instance $X$, its distance to each hyper-plane $H_{i,j}$ is computed. Then $X$ is assigned to class $i$ or $j$. At the end of this process, for every instance $X$, each class $i$ has accumulated a certain number of votes, represented as $V_i(X)$, which is the number of times that the classifier has assigned instance $X$ to class $i$. The final class of $X$, denoted by $C(X)$, will be the one with the highest number of votes.

In the process described above, it is easy to compute the confidence values based on the distance measures of equation 1 (i.e., the closer a case is to the support vectors, the less it is confident). More precisely, in the multi-class classification, we define the confidence of instance $X$ as the sum of its distances to all its class-separating hyper-planes:

$$\varphi(X) = \left| \sum_{H \in \mathcal{H}_{C(x)}} d(X, H) \right| \tag{2}$$

Based on equation 2, the larger value of $\varphi(X)$ shows that $X$ is more confident, and vice versa.

**Step 5**: in this step, BCDC chooses the $K$ most confident temporal relations of those detected in step 4.

**Step 6**: we then retrain the SVM by injecting the temporal relations selected in step 5. It should be noted that for each test document, the original model is retrained by the most confident relations from the documents related to only that test document and not any other test documents.

the model is trained on only the original training data plus the most confident predicted relations from the "relevant" documents for the current test document and not any of predicted relations for other test documents.

Steps 4-6 are repeated until one of the following two termination conditions will be satisfied: 1) there will be no more unselected temporal relation, or 2) a predefined number of iterations will be reached.

**Step 7**: when the retraining phase of the general model for a selected test document is finished, the temporal relations of the test document are classified based on the new specifically retrained model.

Then, if there are still some unprocessed test documents, BCDC will start from step 1 again; otherwise the algorithm will terminate.

The fundamental idea of the second stage of BCDC is to obtain some document- and cluster-wide statistics about the temporal relations between different types of events, and then using this information to improve temporal relation identification.

As it was explained above, a specific model is learned for each test document, using a number of unannotated text documents which are topically related to that test document (i.e., bootstrapping phase). However, if some test documents are themselves topically related, their corresponding retrained models will be very similar. For the sake of efficiency, we can run the bootstrapping phase for just one of such test documents, and then use the same retrained model for the rest. In other words, we run steps 2 to 6 of BCDC just for one member of a set of similar test documents, and for other members, we solely apply step 7.

As it was explained in the section 3.1, we do not explicitly use the assumption of *"one type of temporal relation per discourse"* in any part of BCDC. However, in bootstrapping, we somehow implicitly benefit from this assumption by seeking only topically related documents, which are more likely to include similar event pairs with identical temporal relations.

## 4. Experimental Results of BCDC

In this section, the specification of the employed corpora is briefly explained. Then, the accuracy of BCDC is analyzed.

### 4.1 Characteristic of Corpora

We have used two standard corpora (i.e., TimeBank (v 1.2) and Opinion, see Mani et al., 2006) in our experiments. TimeBank has 183 newswire documents with $64,077$ tokens, and Opinion has 73 documents with 38, 709 tokens. These two datasets have been annotated based on the TimeML standard (Pustejovsky et al., 2003). As mentioned before, there are fourteen temporal relation types (*SIMULTANEOUS*, *IDENTITY*, *BEFORE*, *AFTER*, *IBEFORE*, *IAFTER*, *INCLUDES*, *IS_INCLUDED*, *DURING*, *DURING_INV*, *BEGINS*, *BEGUN_BY*, *ENDS*, *ENDED_BY*) in the TLink class of TimeML. For the sake of reducing the data sparseness problem, as many others (Mani et al., 2006; Tatu & Srikanth, 2008; Mani et al., 2007; Chambers et al., 2007), we have used a normalized version of these relation types including only six following relations:

| | | |
|---|---|---|
| *SIMULTANEOUS* | *ENDS* | *BEGINS* |
| *BEFORE* | *IBEFORE* | *INCLUDES* |

| Original Relation | Converted Relation |
|---|---|
| X *AFTER* Y | Y *BEFORE* X |
| X *IAFTER* Y | Y *IBEFORE* X |
| X *ENDED_BY* Y | Y *ENDS* X |
| X *BEGUN_BY* Y | Y *BEGINS* X |
| X *IS_INCLUDED* Y | Y *INCLUDES* X |
| X *DURING* Y | Y *INCLUDES* X |
| X *IDENTITY* Y | X *SIMULTANEOUS* Y |
| X *DURING_INV* Y | X *INCLUDES* Y |

Table 3: The normalization process for temporal relation types.

For normalizing, the inverse relations are merged. These conversions are shown in the Table 3. In the first six conversions, relations can be easily converted by swapping their arguments. Relations *IDENTITY* and *SIMULTAENOUS* are collapsed, since *IDENTITY* is a subtype of *SIMULTANEOUS* (i.e., two events are *IDENTITY* if they are *SIMULTA-NEOUS* and coreferential). Similarly, relations *DURING_INV* and *INCLUDES* are also collapsed because *DURING_INV* is a subtype of *INCLUDES* (i.e., identical to the Allen's *CONTAINS*) based on the Allen's interval algebra (Allen, 1984). It should be clear that by using these conversions, no information is lost.

| Relation Type | TimeBank | OTC |
|---|---|---|
| *IBEFORE* | 63 | 131 |
| *BEGINS* | 77 | 160 |
| *ENDS* | 114 | 208 |
| *SIMULTANEOUS* | 1304 | 1528 |
| *INCLUDES* | 588 | 950 |
| *BEFORE* | 1335 (38.35%) | 3170 (51.57%) |
| **TOTAL** | 3481 | 6147 |

Table 4: The normalized TLink class distribution in TimeBank and OTC.

In our experiments, like some previous work (Mani et al., 2006; Chambers et al., 2007; Chambers & Jurafsky, 2008), TimeBank and Opinion corpora have been merged into a single corpus called Opinion TimeBank Corpus (OTC). Table 4 shows the normalized TLink class distribution (only for Event-Event relations) over TimeBank and OTC. As it is shown, relation *"BEFORE"* is the most frequent relation; thus it forms the majority class, and can be used as a baseline of the experiments.

For comparison with some other methods, we also used the English part of the TempEval-2 corpus. This part is based on TimeBank (Verhagen et al., 2010; Pustejovsky et al., 2003; Boguraev, Pustejovsky, Ando, & Verhagen, 2007). However, all the TimeBank annotations have been reviewed based on the guidelines of TempEval 2010 and the temporal relations have been modified according to the specific types of the shared task.

| Relation Type | Task E | | Task F | |
|---|---|---|---|---|
| | Training | Test | Training | Test |
| *BEFORE* | 403 | 58 | 600 (35.46%) | 92 |
| *AFTER* | 279 | 38 | 299 | 45 |
| *OVERLAP* | 652 (41.19%) | 124 (48.63%) | 518 | 100 (33%) |
| *BEFORE-OR-OVERLAP* | 61 | 9 | 111 | 55 |
| *OVERLAP-OR-AFTER* | 51 | 6 | 89 | 11 |
| *VAGUE* | 137 | 20 | 75 | 0 |
| **TOTAL** | 1583 | 255 | 1692 | 303 |

Table 5: The distribution of temporal relation types in TempEval-2 Corpus for Task E and F.

There are two parts in this corpus: 1) the training part including 163 documents and $53,450$ tokens; and 2) the test part with 21 documents and $4,848$ tokens. There are six different temporal relation types: *BEFORE*, *AFTER*, *OVERLAP*, *BEFORE-OR-OVERLAP*, *OVERLAP-OR-AFTER*, and *VAGUE*. Among six different tasks of TempEval 2010, we just focused on tasks E and F, which are similar to the problem that we have tackled in this paper. Tasks E and F are the only tasks which consider exclusively the relations between two events. The distribution of temporal relation types for these tasks over the training and test parts of the corpus is shown in Table 5. The majority classes have been underlined in the table.

As it was discussed in section 3.3.2, for each text, we retrieve a number of topically related texts using a public domain software called INDRI. In our experiments, these related texts have been retrieved from the English part of TDT5 multilingual news text corpus[2]. In total, TDT5 consists of $407,505$ text documents in English ($278,109$ documents), Mandarin Chinese ($56,486$ documents), and modern standard Arabic ($72,910$ documents). It also has 250 different topics. Unlike previous TDT corpora, TDT5 does not contain any broadcast news data; all sources are newswires.

---

2. TDT 2004: Annotation Manual, Available at http://www.ldc.upenn.edu/Projects/TDT2004.

## 4.2 Experiments

We have used the LIBSVM java source for the SVM classification (Chang & Lin, 2011). The EVITA system (Saurí et al., 2005) has been used for event extraction. EVITA works based on both linguistic and statistical information. In addition to event extraction, event attributes (which were described in Table 2) can also be extracted by EVITA. We have also used the Stanford NLP package[3] for tokenization, sentence segmentation, part of speech tagging, and parsing. The INDRI retrieval system (Strohman et al., 2005) has been employed to obtain related documents. INDRI is a language model based search engine that provides a state-of-the-art text search engine. The English part of TDT5 has been indexed by INDRI, and by using this search engine, the texts that are highly related to some specified documents can be retrieved.

As it was mentioned earlier, we applied our algorithm to TimeBank, OTC, and TempEval 2010 Corpora. We randomly selected 20 documents (almost 10 percent of total documents) of TimeBank as our development set. Based on several experiments on this development set and with different number of extracted related documents in step 2 of BCDC (i.e., N), and number of most confident relations chosen in step 5 (i.e., $K$), we have set $N$ to 25 and $K$ to 40.

On TimeBank and OTC, the results were evaluated by first excluding the 20 documents of the development set and then measuring accuracy using the five-fold cross validation method. However, for the corpus of TempEval-2, there was no need for cross validation, because the training and test sets are predetermined, and we just reported the accuracy of BCDC on the test set.

Table 6 shows the results of three different settings of the proposed algorithm against several others over TimeBank and OTC. In this table, the baseline is the majority class for event-event relations (i.e., the *BEFORE* relation) of the evaluated corpora. Mani's method is regarded as a successful statistical approach to temporal relation identification, which exclusively uses gold standard features of events (Mani et al., 2007). Methods proposed by Chambers and Mani are similar except that Chambers has also used a number of extra features in a two step algorithm. His method is currently regarded as the state-of-the-art of statistical approaches over TimeBank and OTC. To achieve a higher accuracy, he has also used some extra resources such as WordNet (Chambers et al., 2007).

Argument ancestor path distance (AAPD) is an accurate convolution tree kernel which only uses parse trees of event-event sentences for temporal relation classification (Mirroshandel et al., 2009b). AAPD polynomial is a composite kernel that combines a simple event kernel and AAPD (Mirroshandel et al., 2009b). The mentioned simple event kernel is a linear kernel that exclusively uses the same features as that of Mani's method (Mirroshandel et al., 2009b). AAPD and AAPD polynomial kernels were designed to be applied only to the event pairs that are within the same sentence. Accordingly, the relations of TimeBank and OTC were split into two parts: 1) relations between intra-sentential event pairs, and 2) relations between inter-sentential event pairs. Then these kernels were applied only to the first part and for the second part, we just used simple event kernel (i.e., Mani's kernel). In Table 6, the results reported for AAPD and AAPD polynomial kernel are in fact the outcome of merging the partial results from these two parts.

---

3. Available at http://nlp.stanford.edu/software/index.shtml

| Method | TimeBank Corpus | OTC Corpus |
|---|---|---|
| *Baseline* | 38.35 | 51.57 |
| *Chambers* | <u>59.43</u> | <u>65.48</u> |
| *Mani (Event Kernel + Basic Features)* | 50.97 | 62.5 |
| *Classic Bootstrapping + Event Kernel + Basic Features* | 53.21 | 63.12 |
| *BCDC + Event Kernel + Basic Features* | 59.71 | 65.19 |
| *AAPD Kernel + Extra Event-Event Features* | 54 | 63.44 |
| *Classic Bootstrapping + AAPD Kernel + Extra Event-Event Features* | 57.98 | 64.53 |
| *BCDC + AAPD Kernel + Extra Event-Event Features* | 62.56 | 66.29 |
| *AAPD Polynomial Kernel + Basic Features + Extra Event-Event Features* | 57.02 | 65.95 |
| *Classic Bootstrapping + AAPD Polynomial Kernel + Basic Features + Extra Event-Event Features* | 59.83 | 66.55 |
| *BCDC + AAPD Polynomial Kernel + Basic Features + Extra Event-Event Features* | **66.18** | **68.07** |

Table 6: The accuracy of proposed methods on event-event temporal relation classification over TimeBank and OTC.

*"BCDC + Event Kernel + Basic Features"* is our bootstrapped algorithm, which only uses basic features, mentioned in section 3.2.1, by applying a simple event kernel (Mirroshandel et al., 2011). In *"BCDC + AAPD Kernel + Extra Event-Event Features"*, we utilized extra event-event features in AAPD kernel. Third setting ( *"BCDC + AAPD Polynomial Kernel + Basic Features +Extra Event-Event Features"*) uses AAPD Polynomial kernel to combine all basic and extra event-event features.

For better comparison, we also applied a classic bootstrapping method with the same SVM kernels and features as that of BCDC. For reporting these results, the initial model was trained on a standard corpus (i.e., like stage one of BCDC). Then, in an iterative manner, most confident samples of all documents (rather than just related documents) were used to retrain the model. Note that in this case, there is no need to the process of retrieving related documents, and only one model is learned for all test documents. In order to find the best value for $K$ (i.e., number of most confident samples) in the classic bootstrapping method, we performed several different experiments on the mentioned development set. Incidentally, our experiments showed that here, too, $K$ should be set to 40.

As Table 6 indicates, *"BCDC + AAPD Kernel + Extra Event-Event Features"* and *"BCDC + AAPD Polynomial Kernel + Basic Features + Extra Event-Event Features"* both show a significant improvement over the state-of-the-art method (i.e., Chambers' method). Comparison between BCDC and classical bootstrapping shows the effectiveness of the proposed idea of extracting the retraining samples only from related documents.

The improvement over TimeBank is more considerable than that of OTC. It seems that different distributions of temporal relations in the two corpora has caused the difference between these improvements. As it is shown in Table 4, in OTC, the majority class (i.e., *BEFORE* relation) has a larger part of the whole corpus. This causes the learning algorithm to become biased towards the *BEFORE* relation, and thus the correct prediction of other relations becomes harder. On the contrary, in TimeBank, the distribution is less biased and thus BCDC has shown more improvement on this corpus.

| Method 1 | Method 2 | P-Value on Time-Bank | P-Value on OTC |
|---|---|---|---|
| BCDC + Event Kernel + Basic Features | Mani (Event Kernel + Basic Features) | 0.0134∗ | 0.0476∗ |
| BCDC + AAPD Kernel + Extra Event-Event Features | AAPD Kernel + Extra Event-Event Features | 0.0296∗ | 0.0383∗ |
| BCDC + AAPD Polynomial Kernel + Basic Features + Extra Event-Event Features | AAPD Polynomial Kernel + Basic Features + Extra Event-Event Features | 0.0212∗ | 0.0422∗ |
| BCDC + Event Kernel + Basic Features | Classic Bootstrapping + Event Kernel + Basic Features | 0.0311∗ | 0.0514 |
| BCDC + AAPD Kernel + Extra Event-Event Features | Classic Bootstrapping + AAPD Kernel + Extra Event-Event Features | 0.0299∗ | 0.0441∗ |
| BCDC + AAPD Polynomial Kernel + Basic Features + Extra Event-Event Features | Classic Bootstrapping + AAPD Polynomial Kernel + Basic Features + Extra Event-Event Features | 0.0402∗ | 0.0491∗ |
| BCDC + Event Kernel + Basic Features | Chambers | 0.0765 | 0.0878 |
| BCDC + AAPD Kernel + Extra Event-Event Features | Chambers | 0.0503 | 0.0487∗ |
| BCDC + AAPD Polynomial Kernel + Basic Features + Extra Event-Event Features | Chambers | 0.0431∗ | 0.0397∗ |

Table 7: The statistical significance test results on our proposed methods.

For testing statistical significance, we applied a type of *"stratified shuffling"*, which is a kind of *"compute-intensive randomized test"*. The null hypothesis (i.e., the two models that produced the observed results are the same) was tested by randomly shuffling the generated output for each event pair between the two models and then re-computing the evaluation metrics (i.e., accuracy in this case). If the difference in a particular metric after a shuffling is equal to or greater than the original observed difference in that metric, then a counter ($nc$) for that metric is incremented. Ideally, we should perform all $2^n$ possible shuffles, where $n$ shows the number of test cases (i.e., event pairs). But, in our case, this is impractical because $n$ is a rather large number. Therefore, as many others, we have tried only 10,000 iterations ($nt$). After finishing all iterations, the p-value (likelihood of incorrectly rejecting the null hypothesis) is simply calculated by $(nc+1)/(nt+1)$ (Yeh, 2000). Table 7 shows the result of the significance test on our proposed methods. In this test, each proposed method was compared with its relevant method.

As it is shown in Table 7, majority of the methods passed the test (the p-value is less than 0.05). There are only four exceptions in which the p-value is slightly greater than 0.05.

Table 8 shows the accuracy of BCDC on the English part of the corpus used in TempEval 2010 for tasks E and F. JU-CSE, NCSU-indi, NCSU-joint, TIPSem, TIPSem-B, TRIOS, and TRIPS are participants of the TempEval 2010 shared task (Verhagen et al., 2010). The other methods are the same as in Table 6. AAPD kernel can only be applied to event pairs within the same sentence. Therefore, we are unable to apply it to task E. For inter-sentential event pairs, AAPD Polynomial kernel is almost similar to simple event kernel, because it cannot use the syntactic parse trees, which are appropriate sources of information.

| Method | Task E | Task F |
|---|---|---|
| *Baseline* | 49 | 33 |
| *JU-CSE* | 56 | 56 |
| *NCSU-indi* | 48 | **66** |
| *NCSU-joint* | 51 | 25 |
| *TIPSem* | 55 | 59 |
| *TIPSem-B* | 55 | 60 |
| *TRIOS* | 56 | 60 |
| *TRIPS* | **58** | 59 |
| *Event Kernel + Basic Features* | 38.02 | 33.23 |
| *BCDC + Event Kernel + Basic Features* | 44.35 | 38.44 |
| *AAPD Kernel + Extra Event-Event Features* | – | 40.71 |
| *BCDC + AAPD Kernel + Extra Event-Event Features* | – | 47.20 |
| *AAPD Polynomial Kernel + Basic Features + Extra Event-Event Features* | 38.54 | 43.51 |
| *BCDC + AAPD Polynomial Kernel + Basic Features + Extra Event-Event Features* | <u>45.62</u> | <u>50.41</u> |

Table 8: The accuracy of proposed methods on tasks E and F of TempEval 2010 shared task.

As it can be seen in Table 8, although BCDC has shown some improvement in the accuracy of temporal relation identification (i.e., in comparison with our base methods), it is generally weaker than almost all the participants of TempEval 2010. We think this weakness is due to the restricted feature set that have been used in BCDC. In other words, majority of participants of TempEval 2010 have used richer feature sets of different levels (e.g., lexical, syntactic, and semantic), while we have just used simple event features (plus a few syntactic features only for task F). We think but have not verified yet that with a richer set of features, BCDC will produce more successful results on TempEval's tasks, too. Besides, we think replacing our base method for a more successful method such as TipSem or TRIPS, can make BCDC competitive with participants of TempEval 2010. However, to show this, we first need to find an appropriate confidence measure for step 4 of BCDC[4], which requires further investigation and is one of our directions in future research.

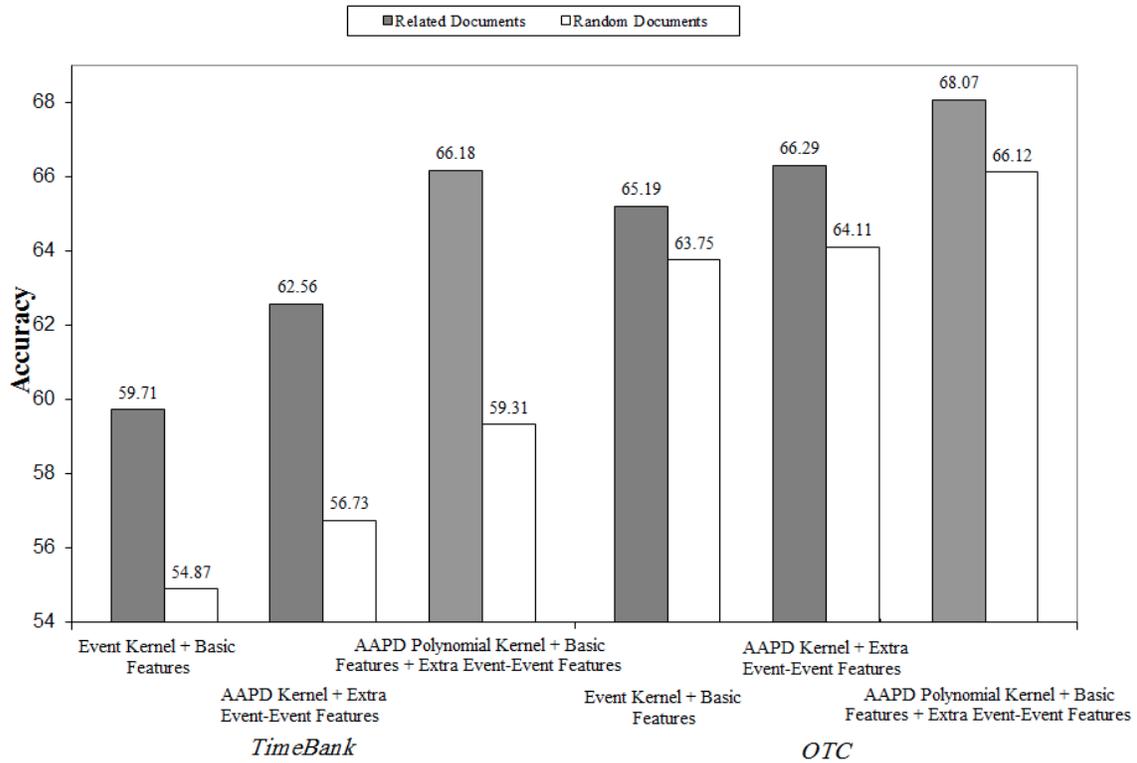

Figure 5: The effects of utilizing related documents vs. randomly selected documents on accuracy of BCDC over TimeBank and OTC.

## 4.3 Analysis

As it can be seen in Tables 6 and 8, BCDC has shown a substantial improvement over several different methods in terms of accuracy and without using any extra annotated data. Bootstrapping by using a number of related documents have the following positive effects:

1) By knowing the relation between events, we can better predict the relation types between analogous events, which may appear in related documents.

2) In related documents, the number of sentences with similar patterns will increase, and the tree kernels can extract more confident information from the parse trees. Thus in this way, SVM can be more informative.

---

4. It should be noted that our proposed confidence measure is just useful for SVM classification technique.

3) The used corpora are rather small with few examples for each relation. This data sparseness problem can affect the performance of any temporal relation identification method. In BCDC, retrieving related documents and extraction of new temporal relations of these documents can increase the number of relations and improve its performance by alleviating the data sparseness problem.

One remaining question is *"what is the impact of choosing related documents?"*. In other words, what if we randomly choose a number of unrelated documents in the bootstrapping phase of BCDC. To show the effectiveness of the idea of using related documents, we have repeated our experiments with $N = 25$ (i.e., the same as original BCDC) randomly selected documents. The results of these experiments are shown in Figure 5. As it is shown, although randomly selected documents have slightly improved the base methods, however, the improvement is not comparable with that of using related documents.

## 5. Using EM for Temporal Relation Learning (EMTRL)

Since supervised and even semi-supervised methods need annotated corpora, which for many languages and/or domains do not exist, here, we propose an unsupervised algorithm for the temporal relation learning problem. Due to the encouraging results of the expectation maximization (EM) algorithm in other unsupervised tasks of natural language processing such as unsupervised grammar induction (Klein, 2005), unsupervised anaphora resolution (Cherry & Bergsma, 2005; Charniak & Elsner, 2009), and unsupervised coreference resolution (Ng, 2008), we decided to evaluate EM in unsupervised temporal relation extraction. Currently, there is no reported work in temporal relation extraction based on EM. In fact, there has not yet been any attempt towards an unsupervised approach to temporal relation extraction. Here, we explain how EM can be successfully applied to the task of temporal relation extraction and show that the performance of EM is encouraging in this task. Before that, we first introduce the definitions and notations that will be later used in subsequent sections.

### 5.1 The EM Algorithm

EM is a general algorithm for maximum likelihood estimation (MLE) (Dempster, Laird, & Rubin, 1977). This algorithm can be used when we deal with incomplete information. As it was mentioned before, in temporal relation learning, the task is to determine the type of temporal relation $r$ that is between two events $e_1$ and $e_2$. In this algorithm, context means the sentence (or sentences) containing a pair of events.

### 5.2 The Proposed Model

Let us call the new proposed algorithm EMTRL, which stands for EM based temporal relation learning. EMTRL operates at the corpus level, inducing valid temporal clustering for all event pairs of a given corpus. More specifically, EMTRL induces a probability distribution to maximize $P(corpus)$ (the probability of the corpus). To easily incorporate linguistic constraints, corpus is represented by its event pairs ($e_i$ $e_j$). We assume event pairs are independent:

$$P(corpus) = \prod_{e_i \ e_j \ \in \ corpus} P(e_i \ e_j) \tag{3}$$

We can rewrite $P(e_i \ e_j)$ so that it uses a hidden variable $TC_{i \ j}$ (temporal class for event pair $e_i \ e_j$) that influences the observed variables ($e_i \ e_j$):

$$P(e_i \ e_j) = \sum_{TC_{i \ j} \ \in \ possible \ temporal \ classes} P(e_i \ e_j, \ TC_{i \ j}) \tag{4}$$

The probability $P(e_i \ e_j, \ TC_{i \ j})$ can be rewritten as:

$$P(e_i \ e_j, \ TC_{i \ j}) = P(e_i \ e_j \mid TC_{i \ j})P(TC_{i \ j}) \tag{5}$$

For inducing temporal relations, EMTRL runs EM on this model. We use a uniform distribution over $P(TC_{i \ j})$. It is clear that if we could choose a more informative prior distribution $P(TC_{i \ j})$, it would have some benefits like having a better handle on the skewness of the distribution. In some other applications of EM, there are settings for this prior distribution. However, in the problem of temporal relation learning, there cannot be any other prior distribution except uniform distribution; because here, all temporal relation types would seem equal to the learner.

If we expand equation 5, each pair $e_i \ e_j$ can be represented by its features, which can be potentially used for determining the temporal relation type between events $e_i$ and $e_j$. Therefore, $P(e_i \ e_j \mid TC_{i \ j})$ can be rewritten using equation 6:

$$P(e_i \ e_j \mid TC_{i \ j}) = P(e_i \ e_j^1, \ e_i \ e_j^2, \ ... \ e_i \ e_j^k \mid TC_{i \ j}) \tag{6}$$

where $e_i \ e_j^l$ is the value of the $l^{th}$ feature of $e_i \ e_j$. These features, which are similar to those mentioned in the work of Chambers and Jurafsky (2008), are listed in Table 9.

To reduce the data sparseness problem and improve the probability estimation, the conditional independence is assumed for these features' value generation. We only assume that tense and aspect are dependent (i.e., $tense_i$ and $aspect_i$), because $tense$ and $aspect$ define temporal location and event structure, and thus considering these features together can be a rich source of information in any temporal relation extraction system. By conditional independence assumption, the value of $P(e_i \ e_j \mid TC_{i \ j})$ can be rewritten as:

$$P(e_i \ e_j \mid TC_{i \ j}) = \prod_{all \ features \ l} P(e_i \ e_j^l \mid TC_{i \ j}) \tag{7}$$

These probabilities (i.e. $P(e_i \ e_j^l \mid TC_{i \ j})$) are regarded as the parameters of our proposed model. Because using them, the likelihood of different temporal classes can be determined. Based on the features in Table 9 and different temporal classes, $P(e_i \ e_j^l \mid TC_{i \ j})$ can be defined. Four examples of such probabilities are shown below:

- $P(class(e_i) = "OCCURRENCE" \ AND \ class(e_j) = "PERCEPTION" \mid TC_{i \ j} = "BEFORE")$

| Feature | Description |
|---|---|
| $Word_1$ & $Word_2$ | The text of first and second events |
| $Lemma_1$ & $Lemma_2$ | The lemmatized first and second events heads |
| $Synset_1$ & $Synset_2$ | The WordNet synset for first and second events heads |
| $POS_1$ & $POS_2$ | The POS of the first and second events |
| Event Government $Verb_1$ & $Verb_2$ | The verbs that govern the first and second events |
| Event Government $Verb_1$ & $Verb_2$ POS | The verbs' POS that govern the first and second events |
| Auxiliary | Any auxiliary adverbs and verbs that modifies the governing verbs |
| $Class_1$ & $Class_2$ | The Class of the first and second events |
| $Tense_1$ & $Tense_2$ | The tense of the first and second events |
| $Aspect_1$ & $Aspect_2$ | The aspect of the first and second events |
| $Modality_1$ & $Modality_2$ | The modality of the first and second events |
| $Polarity_1$ & $Polarity_2$ | The polarity of the first and second events |
| Tense Match | If two events have the same tense or not |
| Aspect Match | If two events have the same aspect or not |
| Class Match | If two events have the same class or not |
| Tense Pair | Pair of two events' tense |
| Aspect Pair | Pair of two events' aspect |
| Class Pair | Pair of two events' class |
| POS pair | Pair of two events' POS |
| $Preposition_1$ | If first event is in a prepositional phrase or not |
| $Preposition_2$ | If second event is in a prepositional phrase or not |
| Text order | If the first event occurs first in the document or not |
| Dominates | If the first event syntactically dominates second event or not |
| Entity Match | If an entity as an argument is shared between two events |

Table 9: The features of events which are used in EMTRL for temporal relation learning.

- $P(e_i \ dominates \ e_j \mid TC_{i \ j} = \text{``}AFTER\text{''})$

- $P(Tense(e_i) = \text{``}PAST\text{''} \ AND \ Tense(e_j) = \text{``}PAST\text{''} \ AND \ Aspect(e_i) = \text{``}NONE\text{''} \ AND \ Aspect(e_j) = \text{``}PROGRESSIVE\text{''} \mid TC_{i \ j} = \text{``}OVERLAP\text{''})$

- $P(POS \ of \ e_i = \text{``}V\text{''} \ AND \ POS \ of \ e_j = \text{``}N\text{''} \mid TC_{i \ j} = \text{``}AFTER\text{''})$

## 5.3 The Induction Algorithm

To induce a temporal clustering on a corpus, EM was applied to our proposed model. In EMTRL, the corpus (i.e., event pairs) and the temporal clustering $TC$ are respectively the observed and unobserved (the hidden) random variables. The EM algorithm includes two main steps of expectation (E) and maximization (M), which in our task can be defined in the following way to iteratively estimate the parameters of the model (i.e., $P(e_i \ e_j^l \mid TC_{i \ j})$):

**E-step**: Fix current parameters of the model, and assign a probability, $P(TC_{i \ j} \mid e_i \ e_j)$, to each possible temporal class for event pairs $(e_i \ e_j)$ of the corpus. This probability can be computed by following equation:

$$P(TC_{i \ j} \mid e_i \ e_j) = \frac{P(e_i \ e_j, \ TC_{i \ j})}{P(e_i \ e_j)} \tag{8}$$

We can rewrite equation 8 by using equations 4, 5, 7:

$$P(TC_{i \ j} \mid e_i \ e_j) = \frac{P(TC_{i \ j}) \prod_{all \ features \ l} P(e_i \ e_j^l \mid TC_{i \ j})}{\sum_{TC'_{i \ j} \ \in \ possible \ temporal \ classes} P(TC'_{i \ j}) \prod_{all \ features \ l} P(e_i \ e_j^l \mid TC'_{i \ j})} \tag{9}$$

Using equation 9, for each event pair $(e_i e_j)$, the temporal relation type (temporal class) with the highest probability is selected. These relations will be later used in the M-step to update the parameters of the model.

**M-step**: By fixing determined temporal relations in E-step, the parameters of the model, $P(e_i \ e_j^l \mid TC_{i \ j})$, are updated in this step. For achieving this goal, different optimization algorithms such as conjugate gradient can be used. However, these algorithms are slow and costly. In addition, it is difficult to smooth these methods in a desired manner. Therefore, we have used the relative frequency method for re-estimation of the parameters, using equation 10:

$$P(e_i \ e_j^l \mid TC_{i \ j}) = \frac{N(e_i \ e_j^l, \ TC_{i \ j})}{N(TC_{i \ j})} \tag{10}$$

where $N(*)$ counts the number of times that given items or joint items have appeared in the corpus. For example, updating probability $P(e_i \ dominates \ e_j \mid TC_{i \ j} = \text{``}AFTER\text{''})$ can be done by dividing $N(e_i \ dominates \ e_j, \ TC_{i \ j} = \text{``}AFTER\text{''})$ (i.e., number of times that $e_i$ dominates $e_j$ and the temporal relation between $e_i$ and $e_j$ is $AFTER$) by $N(TC_{i \ j} = \text{``}AFTER\text{''})$ (i.e., number of times that relation between event pairs in the corpus is $AFTER$).

Steps E and M are repeated until one of the following termination conditions will be satisfied: 1) a predefined number of iterations will be reached, or 2) there will be no more changes in $P(e_i\ e_j^l\ |\ TC_{i\ j})$. In practice, EMTRL is usually stopped after 30 predefined iterations, while final behavior had been apparent after $15 - 22$ iterations.

After finishing the training phase, the temporal relation $(T\hat{C}_{i\ j})$ for requested event pairs $e_i\ e_j$ can be determined using the following equation:

$$T\hat{C}_{i\ j} = \arg\ \max_{TC'\ \in\ possible\ temporal\ classes} P(TC'\ |\ e_i\ e_j) \tag{11}$$

Now, the EM algorithm can begin at either the E-Step or the M-step. We start the induction algorithm at the M-step. It is clear that parameters of the model are not available in the first iteration of EM. Instead, an initial distribution over temporal clustering can be used. There is an important question: how one should initialize this distribution?

Initialization is an important task in EM, because EM only guarantees to find a local maximum of likelihood. The quality of such a local maxima is highly dependent on the initial starting point. We tested three different ways of initialization:

**1) Random Initialization:** a uniform distribution over all temporal clustering was used; therefore, all temporal clustering in the first step had equal probability.

**2) 10% Supervised Initialization:** we used a small part of a labeled corpus (10% of each relation type) for this task. Relations were selected randomly.

**3) Rule-based Initialization:** we used specific rules for initial estimation of temporal relation types and used this initial estimation for computing parameters of the model. These rules were the combination of the so-called GTag rules (Mani et al., 2006), VerbOcean (Chklovski & Pantel, 2005), and rules derived from certain signal words (e.g., "on", "during", "when", and "if") of the text. GTag contains 187 syntactic and lexical rules for inferring and labeling temporal relations between event, document time, and time expressions. Out of these rules, 169 are between event pairs, which were utilized in EMTRL. These 169 rules are either between event pairs of the same sentence or between two main events of two consecutive sentences. An example of a GTag rule is shown below; other rules are accessible from the Blinker part of the TARSQI toolkit[5].

**if** $conjBetweenEvents\ =\ YES$ **&&**
    $isTheSameSentence\ =\ TRUE$ **&&**
    $event_1.class\ =\ (OCCURRENCE|PERCEPTION|ASPECTUAL|I\_ACTION)$ **&&**
    $event_2.class\ =\ STATE$ **&&**
    $event_1.tense\ =\ PAST$ **&&**
    $event_2.tense\ =\ PAST$ **&&**
    $event_1.aspect\ =\ NONE$ **&&**
    $event_2.aspect\ =\ PERFECT$ **&&**
    $event_1.pos\ = VERB$ **&&**
    $event_2.pos\ = VERB$

---

5. Available at http://www.timeml.org/site/tarsqi/index.html

**Then**
    $relation(event_1,\ event_2) = AFTER$

VerbOcean contains lexical rules between two verbs, which can be mined using some lexical and syntactic patterns. The relation between verb pairs can be one of different semantic relations such as strength, enablement, antonymy, similarity, and happens-before. We extracted $4,205$ happens-before rules from VerbOcean. Two examples of these rules are shown below:

    *announce [happens-before] postpone :: 12.844086*
    *review [happens-before] recommend :: 9.049530*

Each rule contains two verbs, their relation, and the strength value of the relationship. For example, the second rule shows relation happens-before between *"review"* and *"recommend"* with strength of 9.049530. We also designed 23 other rules based on some signal words such as *"before"*, *"on"*, *"when"*. These rules are in the GTag format. An example of this group of rules is given below:

**if** *isTheSameSentence = True* **&&**
    *signal = before* **&&**
    *signalBetweenTwoEvents = True*
**Then**
    $relation(event_1,\ event_2) = before$

Like many other statistical NLP tasks, smoothing is vital here to alleviate the problem of data sparseness. In particular, in the first few iterations, much more smoothing is required than in later iterations. In our experiments, we used simply the add-1 smoothing technique in computing equation 10.

## 6. Experimental Results of EMTRL

Like our experiments with BCDC, TimeBank and OTC were also used in the experiments with EMTRL. However, in order to simplify the task, we used a different normalized version of these corpora, which contained only the three following temporal relations:

$$BEFORE \quad AFTER \quad OVERLAP$$

The main reason for this simplification in EMTRL was that reducing the level of supervision in the task of temporal relation learning makes it an even more difficult task, which is itself already considered to be a hard one (Mani et al., 2006). To normalize these corpora and reduce the number of relation types to three, we adopted the same normalization approach like some previous work (Bethard et al., 2007b), *BEFORE* and *IBEFORE* relations were merged into only *BEFORE* relations. Similarly, the *AFTER* and *IAFTER* relations should also be merged into *AFTER* relations. All the remaining ten relation types were collapsed in *OVERLAP* relations. Table 10 shows the converted TLink class distribution over TimeBank and OTC.

| Relation Type | TimeBank Corpus | OTC Corpus |
|:---:|:---:|:---:|
| BEFORE | 706 | 2369 |
| AFTER | 692 | 1073 |
| OVERLAP | 2083 (59.83 %) | 2792 (44.79 %) |
| **Total** | **3481** | **6234** |

Table 10: The converted TLink class distribution in TimeBank and OTC.

Beside TimeBank and OTC, the performance of EMTRL has been also evaluated on tasks E and F of TempEval-2 corpus. The tasks and relations distribution are the same as those shown in Table 5.

## 6.1 Results and Discussions

In our experiments, the baselines were the majority class of event pair relations in the employed corpora (i.e., $OVERLAP$ in both corpora). Note that the Mani's method is in fact supervised, which exclusively uses gold-standard features (Mani et al., 2007). The Chambers' method is similar to Mani's, except that it also uses some external resources such as WordNet (Chambers et al., 2007). Here, the result of our implementation of Mani and Chambers methods are different from their reported results, because, as it was explained before, we only considered three temporal relation types while in their reported experiments, there were six relation types.

In Table 11, in addition to the results of employing EMTRL with three different initializations, we have also reported the results of these initializations as stand-alone classifiers. For Random Initialization and EMTRL + Random Initialization, a question that may arise is how these methods can determine the label of different classes. In fact, these methods can only distinguish three different classes ($Class_1$, $Class_2$, and $Class_3$). Among different possible ways that these unlabeled classes can be mapped to $BEFORE$, $AFTER$, or $OVERLAP$, we choose the mapping in which the similarity between predicted and annotated temporal relations is maximized.

| Method Type | TimeBank Corpus | OTC Corpus |
|:---:|:---:|:---:|
| Baseline | 59.83 | 44.79 |
| Mani | 61.55 | 60.58 |
| Chambers | 66.79 | 62.94 |
| Random Initialization | 35.99 | 37.29 |
| EMTRL + Random Initialization | 40.92 | 43.02 |
| 10% Supervised Initialization | 39.33 | 41.14 |
| EMTRL + 10% Supervised Initialization | **48.31** | **49.34** |
| Rule-based Initialization | 38.92 | 41.03 |
| EMTRL + Rule-based Initialization | 47.86 | 48.78 |

Table 11: The accuracy results of different methods on TimeBank and OTC.

Considering the unsupervised nature of EMTRL, the results of Table 11 can be encouraging. As it is shown in the table, TimeBank's baseline is well above that of OTC. That is because TimeBank is highly biased towards OVERLAP. Accordingly, it is more difficult for learning methods to pass the baseline of TimeBank. The performance of the Mani's method, which is a fully supervised approach, is only slightly over this baseline. In this case, EMTRL's accuracy is considerably below the baseline. However, in the case of OTC, its performance has passed the baseline.

As it is shown in Table 11, EMTRL's accuracy in all three different initializations, have been respectively superior to that of the stand-alone counterparts. The statistical significance of all those results in this table that shows the superiority of EMTRL over the baseline (i.e., in the case of OTC) or the stand-alone initializations (i.e., both corpora) have been verified by the stratified shuffling test with significance level $\alpha = 0.05$.

Table 11 shows that the best accuracy belongs to the Chambers' method. However, it should be noted that this method currently has the best-reported results over TimeBank and OTC among all supervised temporal relation extraction methods.

Table 11 also shows that EMTRL + Randomized Initialization has not been efficient in either corpora. It may be due to the fact that randomized initialization in this very hard problem causes some divergence in the probability distribution. On the other hand, two other initializations have shown satisfactory results in tackling the problem. This implies that initialization is a critical factor in EMTRL, and even little source of supervision can be crucial for achieving satisfactory results.

| Method Type | Task E | Task F |
|---|---|---|
| *Baseline* | 49 | 33 |
| *NCSU-indi* | 48 | 66 |
| *TRIPS* | 58 | 59 |
| *Random Initialization* | 24.75 | 19.11 |
| *EMTRL + Random Initialization* | 27.92 | 22.33 |
| *10% Supervised Initialization* | 26.35 | 23.77 |
| *EMTRL + 10% Supervised Initialization* | 32.20 | 26.29 |
| *Rule-based Initialization* | 27.17 | 23.64 |
| *EMTRL+ Rule-based Initialization* | **32.76** | **27.03** |

Table 12: The accuracy results of different methods on tasks E and F over TempEval-2 Corpus.

Table 12 shows the results of applying EMTRL to the corpus of TempEval 2010. In comparison with the accuracy of kernels in Table 8, EMTRL could achieve encouraging results. In this case, EMTRL's accuracy in all three different initializations, have also been respectively superior to that of the stand-alone counterparts. TRIPS and NCSU-indi are the most successful supervised systems in tasks E and F of TempEval 2010, respectively (Verhagen et al., 2010).

## 6.2 Inconsistency Removal

Since in a pair-wise relation learning system, the relation between each pair of events is predicted without considering its impact on the relations of other event pairs, system may encounter some inconsistencies among predicted relations. This may happen after selecting temporal relations by equation 9 in E-step. It can also happen in finding final class labels by equation 11. Figure 6 shows an example of an inconsistent relation between events $A$, $B$, and $C$:

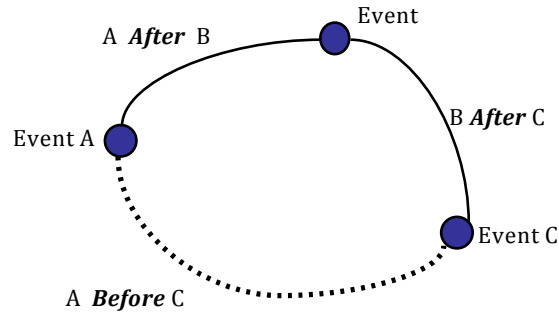

Figure 6: A contradiction in temporal relations between three events A, B, and C.

There are several ways of eliminating such inconsistencies (Mani et al., 2007; Tatu & Srikanth, 2008; Chambers & Jurafsky, 2008). In this work, we have used two different approaches: a greedy best-first search strategy and an Integer Linear Programming (ILP) based method. More details about both approaches are given next.

### 6.2.1 Greedy Best-First Search Strategy

In order to detect possible inconsistencies between predicted relations, we first build a graph for each text, where each node corresponds to an event, and an edge represents a temporal relation between corresponding events. Then, any existing contradiction among connected nodes of each graph can be discovered by applying a set of rules (i.e., 640 rules) based on the Allen's interval algebra (Allen, 1984). As an example, consider the following three rules:

- $before(x, y)$ && $before(y, z) \rightarrow before(x, z)$
- $after(x, y)$ && $before(z, y) \rightarrow after(x, z)$
- $after(x, y)$ && $includes(y, z) \rightarrow after(x, z)$

The inconsistent relations of each graph is stored in a sorted list named $SL$, based on a computed confidence score (i.e., $P(TC_{i\ j} \mid e_i\ e_j)$ of equation 9). Thus, in $SL$, the first and the last elements are the most and the least confident relations, respectively.

The algorithm starts from the first relation of $SL$, and pops off this relation and adds it to another list named $FL$. After adding a new relation to $FL$, the algorithm verifies the consistency among relations of $FL$. If the new relation introduces an inconsistency, it will be replaced by the next confident relation between its corresponding events. This replacement may be repeated until that the new relation will be consistent with other relations existing in $FL$. When there are no more contradictions in $FL$, the algorithm will move the next element of $SL$ to $FL$. These operations are iterated until there will remain no more relations in $SL$. The resultant consistent relations in $FL$ can then be used in subsequent M-step or in the final result of EM.

### 6.2.2 The Integer Linear Programming (ILP)

In our second approach, we cast the task of finding most probable temporal relations as an optimization problem. In contrast with the previous method, this approach, which is based on an integer linear programming (ILP) framework, finds an optimal solution based on the parameters of the model, $P(e_i \ e_j^l \mid TC_{i \ j})$. This method is similar to that of Chambers and Jurafsky (2008).

In this ILP framework, for each event pair $(e_i \ , \ e_j)$, there is a relation type M from $e_i$ to $e_j$ denoted by $TR_{i \ j-M}$. The objective function of the framework is defined as follows:

$$\max \sum_i \sum_{j \ > \ i} \left( \sum_M \ (P_{i \ j-M} \ TR_{i \ j-M} \ + \ P_{j \ i-M} \ TR_{j \ i-M}) \right) \tag{12}$$

where $(P_{i \ j-M}$ (i.e., $P(M \mid e_i \ e_j)$ of equation 8) is the probability of the temporal relation of type $M$ from $e_i$ to $e_j$. This objective function maximizes the sum of probabilities of all temporal relations between event pairs of the input text. There are also three following constraints (i.e., 13, 14, 15) on this objective function:

$$\forall i \ \forall j \ \forall M, \ i > j : \ TR_{i \ j-M}, \ TR_{j \ i-M} \ \in \{0, \ 1\} \tag{13}$$

Constraint 13 implies that each $TR_{i \ j-M}$ variable is either zero or one.

$$\forall i \ \forall j, \ i > j : \ \sum_M (TR_{i \ j-M} \ + \ TR_{j \ i-M}) = 1 \tag{14}$$

Constraint 14 ensures that between each pair of events $(e_i$ and $e_j)$, only one $TR_{i \ j-M}$ variable is set to one, and the rest are set to zero. In other words, it is impossible for a pair of events to have two (or more) relations.

$$TR_{i \ j-M1} \ + \ TR_{j \ k-M2} \ - \ TR_{i \ k-M3} \ \leq \ 1 \tag{15}$$

Constraint 15 guarantees the transitivity conditions among event pairs, wherever relations $TR_{i \ j-M1}$ and $TR_{j \ k-M2}$ entail relation $TR_{i \ k-M3}$. It is obvious that the transitivity constraint is effective only when the event pairs are connected to one another. In a disconnected graph, this constraint has little effect. For example, in Figure 6, by considering this constraint and relations $TR_{A \ B-After}$ and $TR_{B \ C-After}$, $TR_{A \ C-After}$ is the only possible relation between events $A$ and $C$.

After generating the set of all constraints for each document, we can use an ILP solver (SCIP[6]) to solve the problem. One important issue about ILP is that this technique is more effective on dense temporal graphs than in sparse ones.

---

6. This ILP solver which is the fastest existing noncommercial mixed integer programming solver. Available at http://scip.zib.de/

After removing contradictions in the temporal relations, generated (consistent) relations can be easily used in updating the probabilities of the model in the M-step. The results of Tables 11 and 12 are without applying the greedy best-first search or ILP. The accuracy results with the greedy and ILP algorithms over TimeBank and OTC are shown in Table 13. Table 14 shows the accuracy results for tasks E and F over the corpus of TempEval 2010. One question that may arise is how we can enforce the transitivity constraints in EM, when we have only labels $Class_1$, $Class_2$, and $Class_3$, rather than $BEFORE$, $AFTER$, and $OVERLAP$. This problem only happens for the case of EMTRL + Random Initialization, for which we have used a prior assignment of $Class_1 = AFTER$, $Class_2 = BEFORE$, and $Class_3 = OVERLAP$. For the other two initializations (i.e., 10% Supervised and Rule-base), this problem does not occur, because our algorithm starts with the actual class labels $BEFORE$, $AFTER$, and $OVERLAP$.

| Method Type | TimeBank | | | OTC | | |
|---|---|---|---|---|---|---|
| | **Base Method** | **Greedy** | **ILP** | **Base Method** | **Greedy** | **ILP** |
| *EMTRL + Random Initialization* | 40.92 | 41.09 | 41.03 | 43.02 | 42.94 | 43.00 |
| *EMTRL + 10% Supervised Initialization* | **48.31** | 49.54 | 50.34 | **49.34** | **50.52** | 51.44 |
| *EMTRL + Rule-based Initialization* | 47.86 | **50.88** | **52.12** | 48.78 | 49.98 | **51.17** |

Table 13: The accuracy results of applying the greedy best-first search strategy and ILP to TimeBank and OTC.

| Method Type | Task E | | | Task F | | |
|---|---|---|---|---|---|---|
| | **Base Method** | **Greedy** | **ILP** | **Base Method** | **Greedy** | **ILP** |
| *EMTRL + Random Initialization* | 27.92 | 28.03 | 28.04 | 22.33 | 22.32 | 22.30 |
| *EMTRL + 10% Supervised Initialization* | 32.20 | **33.54** | 33.92 | 26.29 | 27.94 | 27.92 |
| *EMTRL + Rule-based Initialization* | **32.76** | 33.49 | **34.12** | **27.03** | **28.36** | **28.55** |

Table 14: The accuracy results of applying the greedy best-first strategy and ILP to tasks E and F of TempEval 2010.

Tables 13 and 14 show the impact of utilizing the greedy best first search and ILP approaches in EMTRL against the base method. By using these strategies, some of the inconsistencies that may exist among predicted temporal relations, are removed (in step E of EMTRL) to make the predicted relations more reliable. As a result, in step M, the parameters of the model will be updated more accurately and thus the accuracy of the whole algorithm will iteratively increase.

The significance of the results depicted in Tables 13 and 14 have been verified by the stratified shuffling with significance level $\alpha = 0.05$. As we had expected, the results of these approaches on *EMTRL + Random Initialization* was not statistically significant. On the other hand, in majority of tests on *EMTRL + 10% Supervised Initialization* and *EMTRL + Rule-based Initialization*, where we compared the output of the greedy and ILP algorithms with that of base method, the statistical significance of the results was verified.

## 7. Conclusion and Future Work

In this paper, we have addressed the problem of temporal relation learning between events, which has been a topic of interest since early days of statistical natural language processing. We have concentrated our efforts to reduce the need to annotated corpora as much as possible. Accordingly, in this paper, two new algorithms, a weakly supervised and an unsupervised, were presented.

The first algorithm was a two-stage weakly supervised approach for classification of temporal relations. In the first stage of the algorithm, a SVM based classifier was trained to learn temporal relations of the corpus. Then, in the second stage of the algorithm, a cross-document bootstrapping technique was employed to iteratively improve the model produced in the first stage. By the idea of bootstrapping, which has been inspired by the hypothesis that we have called *"one type of temporal relation between events per discourse"*, for each test document, some global evidences from a cluster of topically related documents refined local decisions made by the initial model. The results of experiment with this new technique showed a significant improvement in terms of accuracy over related work including the state-of-the-art of the statistical methods.

The second proposed algorithm was a novel model that used the EM algorithm with interval algebra reasoning for temporal relation learning. We compared this work with some of the successful fully supervised methods. Our experiments showed encouraging results, considering the low level of supervision that was provided for the algorithm.

Currently, we are working on finding ways of further improvement of our algorithms, and at the same time trying to reduce the supervision level. In BCDC, by extracting semantic features from related documents, we may be able to improve its performance. Inconsistency removal (i.e. ILP and greedy best first search) algorithms can be also employed in BCDC. Besides, employing the hypothesis of *"one type of temporal relation between events per discourse"* as an explicit constraint can be other possible direction for further research. In EMTRL, one can use other sources of information like narrative information, relations between events and document times, and relations between events and time expressions to build a denser temporal graph. This increases the effectiveness of the greedy best first search and integer linear programming algorithms. We also think, but have not verified yet, that using a richer feature set may further improve the accuracy of EMTRL.

## Acknowledgments

The authors wish to thank the associate editor and the anonymous reviewers for their valuable comments.

## References

Allen, J. (1984). Towards a general theory of action and time. *Artificial intelligence, 23*(2), 123–154.

Bethard, S., & Martin, J. (2007). Cu-tmp: Temporal relation classification using syntactic and semantic features. In *Proceedings of the 4th International Workshop on Semantic Evaluations*, pp. 129–132. Association for Computational Linguistics.

Bethard, S., & Martin, J. (2008). Learning semantic links from a corpus of parallel temporal and causal relations. In *Proceedings of the 46th Annual Meeting of the Association for Computational Linguistics on Human Language Technologies: Short Papers*, pp. 177–180. Association for Computational Linguistics.

Bethard, S., Martin, J., & Klingenstein, S. (2007a). Finding temporal structure in text: Machine learning of syntactic temporal relations. *International Journal of Semantic Computing, 1*(4).

Bethard, S., Martin, J., & Klingenstein, S. (2007b). Timelines from text: Identification of syntactic temporal relations. In *Semantic Computing, 2007. ICSC 2007. International Conference on*, pp. 11–18. IEEE.

Bethard, S. (2007). *Finding event, temporal and causal structure in text: A machine learning approach*. Ph.D. thesis, University of Colorado at Boulder.

Boguraev, B., & Ando, R. (2005). Timeml-compliant text analysis for temporal reasoning. In *Proceedings of IJCAI*, Vol. 5, pp. 997–1003.

Boguraev, B., Pustejovsky, J., Ando, R., & Verhagen, M. (2007). Timebank evolution as a community resource for timeml parsing. *Language Resources and Evaluation, 41*(1), 91–115.

Bramsen, P., Deshpande, P., Lee, Y., & Barzilay, R. (2006). Inducing temporal graphs. In *Proceedings of the 2006 Conference on Empirical Methods in Natural Language Processing*, pp. 189–198. Association for Computational Linguistics.

Chambers, N., & Jurafsky, D. (2008). Jointly combining implicit constraints improves temporal ordering. In *Proceedings of the Conference on Empirical Methods in Natural Language Processing*, pp. 698–706. Association for Computational Linguistics.

Chambers, N., Wang, S., & Jurafsky, D. (2007). Classifying temporal relations between events. In *Proceedings of the 45th Annual Meeting of the ACL on Interactive Poster and Demonstration Sessions*, pp. 173–176. Association for Computational Linguistics.

Chang, C., & Lin, C. (2011). Libsvm: a library for support vector machines. *ACM Transactions on Intelligent Systems and Technology (TIST), 2*(3).

Charniak, E., & Elsner, M. (2009). Em works for pronoun anaphora resolution. In *Proceedings of the 12th Conference of the European Chapter of the Association for Computational Linguistics*, pp. 148–156. Association for Computational Linguistics.

Cheng, Y., Asahara, M., & Matsumoto, Y. (2007). Naist. japan: Temporal relation identification using dependency parsed tree. In *Proceedings of the 4th International Workshop on Semantic Evaluations*, pp. 245–248. Association for Computational Linguistics.

Cherry, C., & Bergsma, S. (2005). An expectation maximization approach to pronoun resolution. In *Proceedings of the Ninth Conference on Computational Natural Language Learning*, pp. 88–95. Association for Computational Linguistics.

Chklovski, T., & Pantel, P. (2005). Global path-based refinement of noisy graphs applied to verb semantics. In *Natural Language Processing–IJCNLP 2005*, pp. 792–803. Springer.

Collins, M., & Duffy, N. (2001). Convolution kernels for natural language. In *Proceedings of NIPS*, Vol. 14, pp. 625–632.

Dempster, A., Laird, N., & Rubin, D. (1977). Maximum likelihood from incomplete data via the em algorithm. *Journal of the Royal Statistical Society. Series B (Methodological)*, *39*, 1–38.

Denis, P., & Muller, P. (2010). Comparison of different algebras for inducing the temporal structure of texts. In *Proceedings of the 23rd International Conference on Computational Linguistics*, pp. 250–258. Association for Computational Linguistics.

Denis, P., & Muller, P. (2011). Predicting globally-coherent temporal structures from texts via endpoint inference and graph decomposition. In *Twenty-Second International Joint Conference on Artificial Intelligence*.

Derczynski, L., & Gaizauskas, R. (2010). Usfd2: Annotating temporal expresions and tlinks for tempeval-2. In *Proceedings of the 5th International Workshop on Semantic Evaluation*, pp. 337–340. Association for Computational Linguistics.

Ha, E., Baikadi, A., Licata, C., & Lester, J. (2010). Ncsu: Modeling temporal relations with markov logic and lexical ontology. In *Proceedings of the 5th International Workshop on Semantic Evaluation*, pp. 341–344. Association for Computational Linguistics.

Hagège, C., & Tannier, X. (2007). Xrce-t: Xip temporal module for tempeval campaign. In *Proceedings of the fourth international workshop on semantic evaluations (SemEval-2007)*, pp. 492–495.

Hall, M., Frank, E., Holmes, G., Pfahringer, B., Reutemann, P., & Witten, I. (2009). The weka data mining software: an update. *ACM SIGKDD Explorations Newsletter*, *11*(1), 10–18.

Han, J., & Kambert, M. (2006). *Data Mining: Concepts and Techniques* (second edition). San Francisco: Morgan Kaufmann.

Hepple, M., Setzer, A., & Gaizauskas, R. (2007). Usfd: preliminary exploration of features and classifiers for the tempeval-2007 tasks. In *Proceedings of SemEval*, pp. 438–441.

Ji, H., & Grishman, R. (2008). Refining event extraction through cross-document inference. In *Proceedings of the Joint Conference of the 46th Annual Meeting of the ACL*, pp. 254–262. Association for Computational Linguistics.

Klein, D. (2005). *The Unsupervised Learning of Natural Language Structure*. Ph.D. thesis, Department of Computer Science, Stanford University.

Kolya, A., Ekbal, A., & Bandyopadhyay, S. (2010). Ju_cse_temp: A first step towards evaluating events, time expressions and temporal relations. In *Proceedings of the 5th International Workshop on Semantic Evaluation*, pp. 345–350. Association for Computational Linguistics.

Lapata, M., & Lascarides, A. (2006). Learning sentence-internal temporal relations. *Journal of Artificial Intelligence Research*, *27*(1), 85–117.

Lin, D., & Pantel, P. (2001). Dirt: discovery of inference rules from text. In *Proceedings of the seventh ACM SIGKDD international conference on Knowledge discovery and data mining*, pp. 323–328. ACM.

Llorens, H., Saquete, E., & Navarro, B. (2010). Tipsem (english and spanish): Evaluating crfs and semantic roles in tempeval-2. In *Proceedings of the 5th International Workshop on Semantic Evaluation*, pp. 284–291. Association for Computational Linguistics.

Mani, I., Verhagen, M., Wellner, B., Lee, C., & Pustejovsky, J. (2006). Machine learning of temporal relations. In *Proceedings of the 21st International Conference on Computational Linguistics and the 44th annual meeting of the Association for Computational Linguistics*, pp. 753–760. Association for Computational Linguistics.

Mani, I., Wellner, B., Verhagen, M., & Pustejovsky, J. (2007). Three approaches to learning tlinks in timeml. Tech. rep., Technical Report CS-07-268, Brandeis University.

Mikheev, A., Grover, C., & Moens, M. (1998). Description of the ltg system used for muc-7. In *Proceedings of 7th Message Understanding Conference (MUC-7)*. Fairfax, VA.

Min, C., Srikanth, M., & Fowler, A. (2007). Lcc-te: a hybrid approach to temporal relation identification in news text. In *Proceedings of the 4th International Workshop on Semantic Evaluations*, pp. 219–222. Association for Computational Linguistics.

Mirroshandel, S., Ghassem-Sani, G., & Khayyamian, M. (2009a). Event-time temporal relation classification using syntactic tree kernels. In *Proceeding of the 4th Language and Technology Conference*, pp. 300–304.

Mirroshandel, S., Ghassem-Sani, G., & Khayyamian, M. (2009b). Using tree kernels for classifying temporal relations between events. In *Proceedings of the 23th Pacific Asia Conference on Language, Information and Computation*, pp. 355–364.

Mirroshandel, S., Ghassem-Sani, G., & Khayyamian, M. (2011). Using syntactic-based kernels for classifying temporal relations. *Journal of Computer Science and Technology*, *26*(1), 68–80.

Mulkar-Mehta, R., Hobbs, J., Liu, C., & Zhou, X. (2009). Discovering causal and temporal relations in biomedical texts recognizing causal and temporal relations. In *Proceedings of the AAAI Spring Symposium, Stanford CA*.

Muller, P., & Tannier, X. (2004). Annotating and measuring temporal relations in texts. In *Proceedings of the 20th international conference on Computational Linguistics*, pp. 50–56. Association for Computational Linguistics.

Ng, V. (2008). Unsupervised models for coreference resolution. In *Proceedings of the Conference on Empirical Methods in Natural Language Processing*, pp. 640–649. Association for Computational Linguistics.

Pekar, V. (2006). Acquisition of verb entailment from text. In *Proceedings of the main conference on Human Language Technology Conference of the North American Chapter of the Association of Computational Linguistics*, pp. 49–56. Association for Computational Linguistics.

Petrov, S., & Klein, D. (2007). Improved inference for unlexicalized parsing. In *Proceedings of NAACL HLT 2007*, pp. 404–411.

Puscasu, G. (2007). Wvali: Temporal relation identification by syntactico-semantic analysis. In *Proceedings of the 4th International Workshop on SemEval*, pp. 484–487.

Pustejovsky, J., Hanks, P., Sauri, R., See, A., Gaizauskas, R., Setzer, A., Radev, D., Sundheim, B., Day, D., Ferro, L., & Lazo, M. (2003). The timebank corpus. In *Corpus Linguistics*, Vol. 2003, p. 40.

Saurí, R., Knippen, R., Verhagen, M., & Pustejovsky, J. (2005). Evita: a robust event recognizer for qa systems. In *Proceedings of the conference on Human Language Technology and Empirical Methods in Natural Language Processing*, pp. 700–707. Association for Computational Linguistics.

Søgaard, A. (2011). Semisupervised condensed nearest neighbor for part-of-speech tagging. In *Proceedings of the 49th Annual Meeting of the Association for Computational Linguistics: Human Language Technologies: short papers*, Vol. 2, pp. 48–52.

Strohman, T., Metzler, D., Turtle, H., & Croft, W. (2005). Indri: A language model-based search engine for complex queries. In *Proceedings of the International Conference on Intelligent Analysis*.

Szpektor, I., Tanev, H., Dagan, I., & Coppola, B. (2004). Scaling web-based acquisition of entailment relations. In *Proceedings of EMNLP*, Vol. 4, pp. 41–48.

Tatu, M., & Srikanth, M. (2008). Experiments with reasoning for temporal relations between events. In *Proceedings of the 22nd International Conference on Computational Linguistics-Volume 1*, pp. 857–864. Association for Computational Linguistics.

UzZaman, N., & Allen, J. (2010). Trips and trios system for tempeval-2: Extracting temporal information from text. In *Proceedings of the 5th International Workshop on Semantic Evaluation*, pp. 276–283. Association for Computational Linguistics.

Verhagen, M., Gaizauskas, R., Schilder, F., Hepple, M., Katz, G., & Pustejovsky, J. (2007). Semeval-2007 task 15: Tempeval temporal relation identification. In *Proceedings of the 4th International Workshop on Semantic Evaluations*, pp. 75–80. Association for Computational Linguistics.

Verhagen, M., Sauri, R., Caselli, T., & Pustejovsky, J. (2010). Semeval-2010 task 13: Tempeval-2. In *Proceedings of the 5th International Workshop on Semantic Evaluation*, pp. 57–62. Association for Computational Linguistics.

Yarowsky, D. (1995). Unsupervised word sense disambiguation rivaling supervised methods. In *Proceedings of the 33rd annual meeting on Association for Computational Linguistics*, pp. 189–196. Association for Computational Linguistics.

Yeh, A. (2000). More accurate tests for the statistical significance of result differences. In *Proceedings of the 18th conference on Computational linguistics-Volume 2*, pp. 947–953. Association for Computational Linguistics.

Yoshikawa, K., Riedel, S., Asahara, M., & Matsumoto, Y. (2009). Jointly identifying temporal relations with markov logic. In *Proceedings of the Joint Conference of the 47th Annual Meeting of the ACL and the 4th International Joint Conference on Natural Language Processing of the AFNLP: Volume 1-Volume 1*, pp. 405–413. Association for Computational Linguistics.

Zhang, M., Zhang, J., Su, J., & Zhou, G. (2006). A composite kernel to extract relations between entities with both flat and structured features. In *Proceedings of the 21st International Conference on Computational Linguistics and the 44th annual meeting of the Association for Computational Linguistics*, pp. 825–832. Association for Computational Linguistics.